\documentclass[10pt,journal]{IEEEtran}
\usepackage{cite,bm,xspace}
\usepackage{hyperref}
\usepackage{xcolor}
\usepackage{tikz-cd}

\usepackage{amsmath,amssymb,amsfonts}
\usepackage{algorithmic}
\usepackage{graphicx}
\usepackage{textcomp}
\usepackage{dblfloatfix}
\usepackage{multirow}
\usepackage{subcaption}
\usepackage{verbatim}
\def\BibTeX{{\rm B\kern-.05em{\sc i\kern-.025em b}\kern-.08em
    T\kern-.1667em\lower.7ex\hbox{E}\kern-.125emX}}

\long\def\comment#1{} 

\newcommand{\xmath}[1] {\ensuremath{#1}\xspace}
\newcommand{\blmath}[1] {\xmath{\bm{#1}}}

\newcommand{\1}{\blmath{1}} 
\newcommand{\0}{\blmath{0}}

\newcommand{\Ib}{{\blmath I}}

\newcommand{\Xb}{{\blmath X}}

\newcommand{\mb}{{\blmath m}}

\newcommand{\xb}{{\blmath x}}
\newcommand{\yb}{{\blmath y}}
\newcommand{\zb}{{\blmath z}}

\newcommand{\Hc}{\mathcal{H}}
\newcommand{\Mc}{\mathcal{M}}

\newcommand{\Tc}{\mathcal{T}}
\newcommand{\Xc}{\mathcal{X}}
\newcommand{\Yc}{\mathcal{Y}}
\newcommand{\Sc}{\mathcal{S}}
\newcommand{\Sigmab}{{\boldsymbol {\Sigma}}}

\newcommand{\Rd}{{\mathbb R}}

\newcommand{\beq}{\begin{equation}}
\newcommand{\eeq}{\end{equation}}
\newcommand{\beqa}{\begin{eqnarray}}
\newcommand{\eeqa}{\end{eqnarray}}

\begin{document}
\title{Continuous Conversion of CT Kernel  using Switchable CycleGAN with AdaIN}
\author{Serin Yang, Eung Yeop Kim, and Jong Chul Ye, \IEEEmembership{Fellow, IEEE}
\thanks{This work was supported by the National Research Foundation (NRF) of Korea grant NRF-2020R1A2B5B03001980. }
\thanks{S. Yang and J.C. Ye are with the Department of Bio and Brain Engineering, Korea Advanced Institute of Science and Technology (KAIST), Daejeon 34141, Republic of Korea (e-mail: {yangsr, jong.ye}@kaist.ac.kr).  E.Y. Kim is with the Department of Radiology, Samsung Medical Center, Sungkyunkwan University School of Medicine, 81 Irwon-ro Gangnam-gu, Seoul, 06351, Republic of Korea (e-mail: neuroradkim@gmail.com)}
}

\maketitle

\begin{abstract}
X-ray computed tomography (CT) uses different filter kernels to highlight different structures.
Since the raw sinogram data is usually removed after the reconstruction,  in case there are additional need for other types of kernel images that were not previously generated,  the patient may need to be scanned again. Accordingly, there exists increasing demand for post-hoc image domain conversion   from one kernel  to another without sacrificing the image quality. In this paper, we propose a novel unsupervised continuous kernel conversion method using cycle-consistent generative adversarial network (cycleGAN) with adaptive instance normalization (AdaIN).
Even without paired training data,
not only can our network translate the images between two different kernels, but it can also convert images along the interpolation path between the two kernel domains.
We also show that  
the quality of generated images can be further improved 
 if intermediate kernel domain images are available.

Experimental results confirm that our method not only enables accurate kernel conversion that is comparable to supervised learning methods, but also generates intermediate kernel images in the unseen domain that  are useful for hypopharyngeal cancer diagnosis.
\end{abstract}

\begin{IEEEkeywords}
Computed tomography, reconstruction kernels, cycle-consistent adversarial networks, style transfer, adaptive instance normalization (AdaIN)
\end{IEEEkeywords}

\section{Introduction}
\label{sec:introduction}
\IEEEPARstart{I}{n} computed tomography (CT) images, raw sinogram data are collected from detectors, from which
tomographic images are reconstructed using algorithms such as filtered backprojection (FBP).
Depending on the structural property of the object,  a specific reconstruction fitter kernel is selected which affects the range of features that can be seen\cite{setiyono2020optimization}. For example, high pass filters, which are often used 
to bone and tissues with high CT contrast,
 preserve higher spatial frequencies while reducing lower spatial frequencies. On the other hand, low pass filters 
compromise spatial resolution but significantly reduces the noises, which  are adequate for brain or soft tissue imaging \cite{setiyono2020optimization}, \cite{boedeker2004emphysema}.

\begin{figure*}[!t]
	\centerline{\includegraphics[width=2\columnwidth]{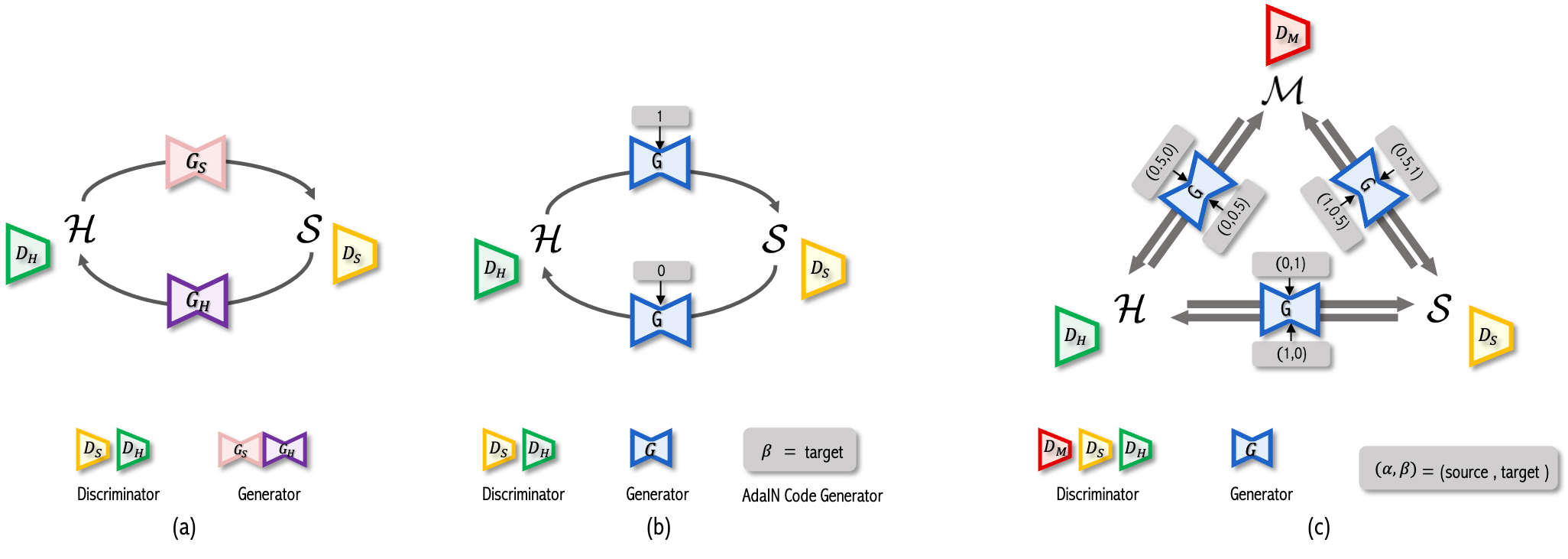}}
	\caption{(a) Vanilla cycleGAN for conversion of two different kernels. (b) Switchable cycleGAN for two domains, and (c) switchable cycleGAN with split AdaIN for three domains.}
	\label{fig_adain_vs_vanilla}
\end{figure*}

Although different kernels would be required in order to examine different structures,
 the size of the required storage can become quickly too large to accommodate various types of kernel images \cite{weiss2011hybrid,takagi2014combined}. Therefore, a routine practice is just to save a particular kernel images. Accordingly, kernel conversion is expected to be useful in case when additional kernel images are required.

For example, in case of  paranasal sinuses CT, only one type of kernel images are usually reconstructed, either sharp or soft kernels. Sharp kernels are applicable to check abnormality of bone but hard to evaluate soft tissue. On the other hand, it is relatively easy to evaluate abnormality in the soft tissue with soft kernels but difficult to see bone details.  Also, in case of temporal bone CT, soft kernel images are occasionally required although sharp kernel images are only acquired in routine clinical practice. In these cases, kernel conversion would be useful for both healthcare experts and patients because it reduces additional burden of extra CT scanning. Through  kernel conversions, both sharp and soft kernel images can be obtained and utilized adequately in various situations.


%
%
Although many researchers have investigated this issue \cite{weiss2011hybrid, takagi2014combined,missert2020synthesizing}, the image domain filter kernel conversion  is still seen as a difficult task since the relationship between different kernel images needs to be found out and a new texture should be synthesized for the target domain.

Recently, 
 a deep learning approach was explored for kernel conversion \cite{lee2019ct}.
This approach is based on the supervised training, so that it can only be applied when there exists paired dataset
from the different kernels.
Although paired data sets for training could be generated from the same sinogram with different filter kernels, collecting all paired CT kernel image data sets for different CT scanners and acquisition conditions like KVp, mAs, etc. would be a daunting task that could only be accomplished by carefully planning the data collection in advance.

Therefore, one of the most important motivations of this work is to use an unsupervised deep learning approach for kernel conversion
that can be trained without paired data set. In particular, we
 consider the kernel conversion problem as
an unsupervised image style transfer problem, and develop an unsupervised deep learning approach.

In fact,
 cycle-consistent adversarial network (cycleGAN) \cite{zhu2017unpaired} is one of the representative unsupervised image style transfer methods that can learn to translate between two different domains.  Furthermore, our recent theoretical work \cite{sim2019optimal} shows that 
 the cycleGAN can be interpreted as an optimal transport between two probabilistic distributions \cite{villani2008optimal,peyre2019computational} that simultaneously
 minimizes the statistical distances between the empirical data and synthesized data in  two domains.
Therefore, we are interested in using cycleGAN as our model, with each domain consisting of  distinct kernel images.

One important contribution is that
unlike the conventional cycleGAN that uses two distinct generators, we employ
the switchable cycleGAN architecture \cite{gu2021adain} for kernel conversion so that
a single conditional generator with
adaptive instance normalization (AdaIN) \cite{huang2017arbitrary} can be used for both forward and backward kernel conversion.
In addition to reducing the memory requirement for the cycleGAN training \cite{gu2021adain},
the proposed cycleGAN generator with AdaIN layer can
generate every  interpolating path along an optimal transport path between two target domains at the
inference phase, which brings us an opportunity to discover new objects or tissues which could not have been observed with the currently given two reconstruction kernels.  

Unfortunately, the existing switchable CycleGAN \cite{gu2021adain} proves difficult to utilize the intermediate kernel images if available. This is because the kernel interpolation is only unidirectional from soft or sharp kernels to the intermediate kernel, but the conversion from the intermediate kernel to the other kernels is not possible.
To remedy this, here we propose a novel {\em split AdaIN} code generator architecture in which two AdaIN code generators are used independently of each other for encoder and decoder. One of the major innovations in this novel architecture is that it now enables bidirectional kernel conversion between any pairs of kernel domains along the interpolation path, thus enabling effective use of the intermediate domain kernel images to greatly improve the overall kernel conversion performance.

This paper is  organized as follows.
CT conversion and image style transfers are first reviewed in Section~\ref{sec:review},
after which we explain how these techniques can be synergistically combined for our kernel conversion algorithm in Section~\ref{sec:theory}.
 Section~\ref{sec:methods} describes the data set and experimental setup. Experimental results are provided in Section~\ref{sec:results}, which is followed
by Discussion and  Conclusions in Section~\ref{sec:discussion} and Section~\ref{sec:conclusion}.

\section{Related works}
\label{sec:review}

\subsection{CT Kernel Conversion}

In classical CT kernel conversion approaches,
one common way is to combine two different kernel images into one for better diagnostic purpose \cite{weiss2011hybrid, takagi2014combined,missert2020synthesizing}.
For example, for hybrid kernel combination, the upper and lower thresholds for pixel intensities are chosen, and then
 if the pixel values of the images reconstructed with low pass filters are outside those thresholds, the values are replaced with those from the high pass filters.
Unfortunately, the optimality of the combined filter kernel is a subjective matter, which  depends on the clinical applications.

In this regard, deep learning approaches for CT kernel conversion \cite{lee2019ct,choe2019deep} do not
interfere with the existing clinical workflows, as the generated images are still in the conventional kernel sets.
Furthermore, a radiomic study revealed that the generated images do not sacrifice the accuracy of the diagnosis \cite{choe2019deep}.
Nonetheless, this method does not generate new kinds of hybrid informations that could be obtained in the aforementioned kernel combination methods.

\subsection{Deep Learning for Image Style Transfer}

Currently, two types of approaches are available for image style transfer. First, a content image and a style reference image are passed to a neural network, and the goal is to convert the content image to imitate styles from the style reference.
For example, Gatys et al.  \cite{gatys2016image} solves an optimization problem in the feature spaces between the content and style images.
However, the iterative optimization process takes significant amount of time and the results are easily overly stylized. 
Instead, the adaptive instance normalization (AdaIN) was proposed as a simple alternative \cite{huang2017arbitrary}. Specifically, AdaIN layer estimates the means and variances of referenced style features and uses them to correct the mean and variance of content features.
Despite the simplicity, a recent theoretical works \cite{mroueh2019wasserstein}
 showed that the style transfer by AdaIN is a special case of optimal transport  \cite{villani2008optimal,peyre2019computational} between two Gaussian distributions.

Another type of style transfer is performed as a distribution matching approach. 
Specifically,   let the target style images lie in the domain $\Xc$
 equipped with a probability measure $\mu$, whereas
the input content images lie in $\Yc$   with a probability measure $\nu$.
Then, the image style transfer is to transport the content distribution $\nu$ to the style image distribution $\mu$, and vice versa.
In our recent theoretical work \cite{sim2019optimal}, we show that this type of style transfer problem can
be solved through  optimal transport  \cite{villani2008optimal,peyre2019computational}.
In particular, if we define the transportation cost as the sum of the statistical distances between the empirical distribution and the generated distribution in $\Xc $ and $\Yc$, respectively, and try to find the joint distribution that minimizes the sum of the distances, then the Kantorovich dual formulation  \cite{villani2008optimal,peyre2019computational} leads to the cycleGAN formulation.
This justifies why cycleGAN has become a representative style transfer method.

\begin{figure}[!t]
	\centerline{\includegraphics[width=\columnwidth]{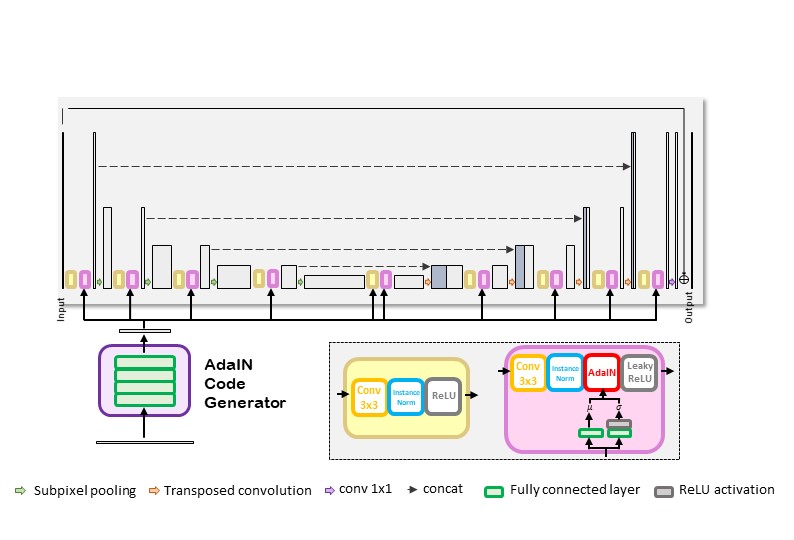}}
	\caption{Generator architecture in two-domain cycleGAN with one AdaIN code generator.}
	\label{fig_2domain_generator_architecture}
\end{figure}

\begin{figure*}[!t]
	\centerline{\includegraphics[width=1.8\columnwidth]{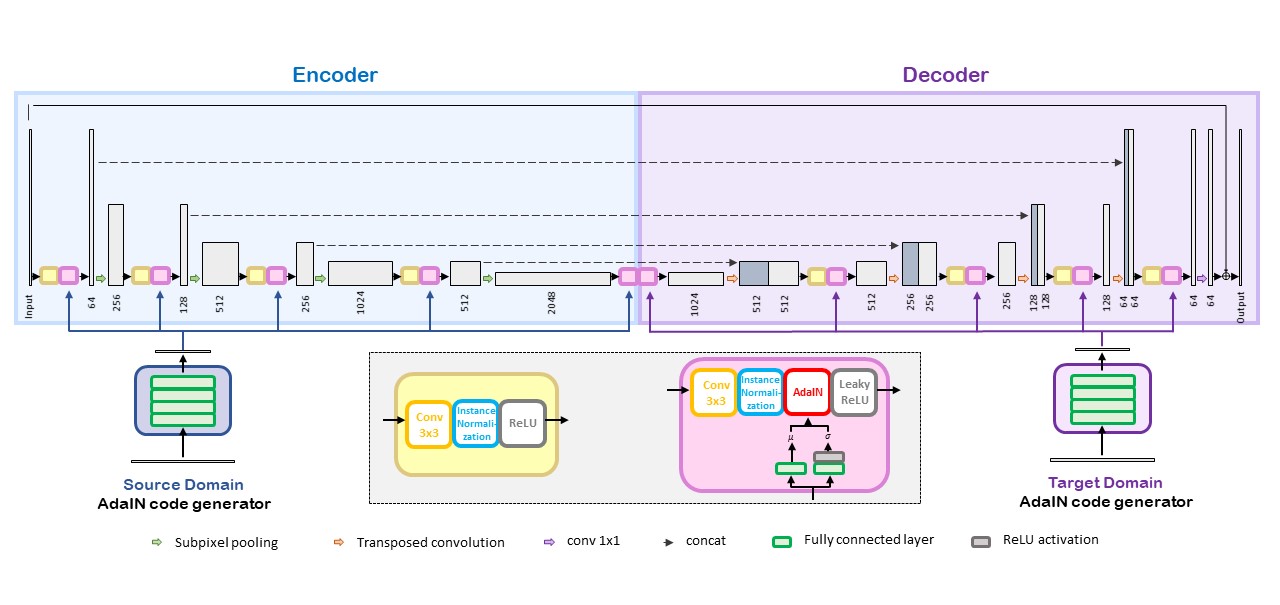}}
	\caption{Generator architecture in multi-domain cycleGAN with two different AdaIN code generators.}
	\label{fig_generator_architecture}
\end{figure*}
 
By synergistically combining the two ideas, we recently proposed
switchable cycleGAN  \cite{gu2021adain}  that combines AdaIN  into cycleGAN so that
only a single generator can be used for style transfer between two domains.
Although the original motivation of  \cite{gu2021adain}  was to reduce the memory requirement for the cycleGAN training by eliminating
additional generator,
the cycleGAN with AdaIN has an important advantage that cannot be achieved using the conventional
cycleGAN. Specifically, the switchable cycleGAN with AdaIN can generate
intermediate images on the interpolating path between two domains in the training data set, leading to diverse
kernel conversions.

Unfortunately, the direct use of switchable cycleGAN turns out difficult to utilize the intermediate kernel images if available. 
As will be described later,
the kernel interpolation by the existing switchable cycleGAN \cite{gu2021adain}  works only unidirectional from soft or sharp kernels to the intermediate kernel, but the conversion from the intermediate kernel data to the other kernel images, which is essential to fully utilize the intermediate kernel images,
 is not possible. This is one of the motivation for the new split AdaIN architecture, which will be also introduced in this paper.

\section{Theory}
\label{sec:theory}

\subsection{Instance Norm, AdaIN and Optimal Transport}

Suppose that a multi-channel feature tensor at a specific layer is represented by
\begin{align}\label{eq:X}
\Xb =&\begin{bmatrix} \xb_1  & \cdots &\xb_P \end{bmatrix} \in \Rd^{HW\times P},
\end{align}
where $P$ is  the number of channel in the feature tensor $\xb$, and
$\xb_i  \in \Rd^{HW\times 1}$ refers to the $i$-th column vector of $\xb$, 
which represents the vectorized feature map of size of $H\times W$  at the $i$-th channel.

Then,  instance normalization  \cite{ulyanov2016instance} and AdaIN \cite{huang2017arbitrary} convert the feature data at each channel using the following transform:
\begin{align}
\zb_i &= \Tc(\xb_i,\yb_i),\quad i=1,\cdots, P
\end{align}
where
\begin{align}\label{eq:AdaIN}
\Tc(\xb,\yb)  := \frac{\sigma(\yb) }{\sigma(\xb) }\left(\xb -m(\xb) \1\right) +m(\yb)\1,\quad 
\end{align}
where $\1 \in \Rd^{HW}$ is the $HW$-dimensional vector composed of 1, and 
 $m(\xb)$ and $\sigma(\xb)$ are the mean and standard deviation of $\xb\in \Rd^{HW}$;
 $m(\yb)$ and $\sigma(\yb)$ refer to the target style domain mean and standard deviation, respectively.

Specifically, the instance norm uses $(\sigma(\yb),m(\yb))=(1,0)$, whereas the
$(\sigma(\yb),m(\yb))$ are computed from the style features for the case of AdaIN.
Eq.~\eqref{eq:AdaIN} implies that the mean and variance of the feature in the content image are
 normalized so that they can match the mean and variance of the style image feature.
In fact, the transform \eqref{eq:AdaIN} is closely related to the optimal transport  between two probabilistic distributions \cite{peyre2019computational,villani2008optimal}.

  Specifically,   let the two probability
spaces $\mathcal{U} \subset \Rd^{HW}$ and $\mathcal{V} \subset \Rd^{HW}$ be
 equipped with the Gaussian probability measure $\mu\sim \mathcal{N}(\mb_U,\Sigmab_U)$  and
$\nu\sim\mathcal{N}(\mb_V,\Sigmab_V)$, respectively, where $\mb_U$ and $\Sigmab_U$ denote the mean
vector and the covariance matrix, respectively.
Then, a  closed form
optimal transport map from the measure $\mu$ to the measure $\nu$ with respect to Wasserstein-2 distance can be obtained as follows \cite{mroueh2019wasserstein}:
\begin{align}\label{eq:T_general}
T_{\mu\rightarrow \nu}(\xb)= \mb_V + \Sigmab_U^{-\frac{1}{2}}\big(\Sigmab_U^{\frac{1}{2}}\Sigmab_V\Sigmab_U^{\frac{1}{2}}\big)^{\frac{1}{2}}\Sigmab_U^{-\frac{1}{2}}(\xb-\mb_U)
\end{align}
In particular, if we assume the independent and identically distributed feature spaces, i.e. 
\begin{align*}
\mb_U=m(\xb)\1, \Sigmab_U=\sigma(\xb)\Ib,  \mb_V=m(\yb), \Sigmab_V=\sigma(\yb)\Ib
\end{align*}
then AdaIN transform  in \eqref{eq:AdaIN} can be obtained as a special case of \eqref{eq:T_general} as follows:
\begin{align}\label{eq:T}
T_{\mu\rightarrow \nu}(\xb)= m(\yb)\1 + \frac{\sigma(\yb)}{\sigma(\xb)}(\xb-m(\xb)\1)
\end{align}
Accordingly, we can also see that instance norm is a special case of optimal transport that transports the feature to the normalized Gaussian distribution with   zero mean and unit variance.

\subsection{Switchable CycleGAN with AdaIN}

Suppose that the domain $\Sc$ is composed of CT images from soft tissue kernel, whereas the images in the domain $\Hc$ are generated
by sharp bone kernel.
Then, as shown in Fig.~\ref{fig_adain_vs_vanilla}(a),  
a standard cycleGAN framework for kernel conversion would require two generators: 
the forward generator from sharp kernel to soft kernel $(G_{S})$, 
the backward generator from soft kernel to sharp kernel images ($G_{H}$).

On the other hand, 
our switchable cycleGAN implements two generators using a single baseline autoencoder network $G$ followed by AdaIN-based optimal transport layers.
Specifically, to generate images in the target domain from source domain inputs,
we use an autoencoder as the baseline network
and then use the AdaIN transform to transport the autoencoder features to the target kernel features.

More specifically, the AdaIN codes  for signaling $\Hc$  and $\Sc$ domains are generated as:
\begin{align}
 \left(m(\yb),\sigma(\yb)\right)=\begin{cases}(1,0), & \mbox{domain $\Hc$} \\ (\sigma_S,m_S), & \mbox{domain $\Sc$} \end{cases}
\end{align} 
Using this, AdaIN code generator $F$ is defined as follows:
\begin{align}\label{eq:F}
F(\beta):=&\begin{bmatrix}\sigma(\beta)\\m(\beta)\end{bmatrix} =  (1-\beta)\begin{bmatrix}1\\0\end{bmatrix} +  \beta\begin{bmatrix}\sigma_S\\ m_S\end{bmatrix}
\end{align}
where $(\sigma_S,m_S)$  are learnable parameters during the training, and $\beta$ is a
variable that represents the domain.

Using this,  we can consider two generator architectures.
First, when the training data are only from two domains such as $\Sc$ and $\Hc$ domains,
 the AdaIN generator
 only needs to specify the target domain, since the source domain is automatically specified.
 For example,  the conversion to the $\Hc$ and $\Sc$ can be done by
\begin{align}
G_H(\xb) &= G_{1,0}(\xb):=G(\xb; F(0))  \label{eq:GH1}\\
G_S(\yb)  &= G_{0,1}(\yb):= G(\yb; F(1)) \label{eq:GS1}  
\end{align}
where the first and second subscripts of the generator $G$ denote the source and target domains, respectively.
Accordingly, this can be implemented
using a single AdaIN code generator  as shown in Fig.~\ref{fig_2domain_generator_architecture}.

On the other hand, if there are additional middle domain data  and we consider
the conversion from them,  we should specify the source domain. In this case, 
the generator should be implemented in the form
\begin{align}\label{eq:Gab}
G_{\alpha,\beta}(\zb):=G(\zb; F_e(\alpha), F_d(\beta))
\end{align}
which implies the conversion from the source in an intermediate domain $\alpha \Hc +(1-\alpha)\Sc$
to a target domain $ \beta\Hc+(1-\beta)\Sc$. Here, $\alpha \Hc +(1-\alpha)\Sc$  represents a domain
that lies between $\Sc$ and $\Hc$ domains, whose conceptual distance is determined by $\alpha$.
For example, the middle kernel domain between $\Sc$ and $\Hc$ can be referred to as $0.5\Hc+0.5\Sc$.

In \eqref{eq:Gab}, to specify the source and target domains, the autoencoder 
is equipped with
 the encoder AdaIN code $F_e(\alpha)$ and decoder code $F_d(\beta)$, which are similar to \eqref{eq:F} but independently implemented.
 Therefore, the split AdaIN code generator
 architecture in Fig.~\ref{fig_generator_architecture} is necessary.
Accordingly, with a slight abuse of notation, 
two generators $G_S$ and $G_H$ can be implemented using a single generator with split AdaIN:
\begin{align}
G_H(\xb) =G_{1,0}(\xb):= G(\xb; F_e(1), F_d(0)) \label{eq:GH}\\
G_S(\yb) = G_{0,1}(\yb):=G(\yb; F_e(0), F_d(1)) \label{eq:GS}
\end{align}
As will become clear,
the split AdaIN   is essential to utilize the intermediate kernel domain data for training, if available.
Moreover, split AdaIN  increases the flexibility so that it allows conversion
between any pairs of  domains along the interpolation path between the $\Hc$ and $\Sc$.

\subsection{Network Training}

\subsubsection{When two domain training data are available}

This corresponds to the scenario in  \cite{gu2021adain}.
In this case, the training can be done by solving the following min-max optimization problem:
\begin{align}\label{eq:minmax}
\min\limits_{G,F}\max\limits_{D_H,D_S}\ell_{total}(G, F, D_S, D_H)
\end{align}
where the total loss is given by
\begin{equation} \label{eq:loss_total}
\begin{split}
&\ell_{total}(G, F, D_S, D_H) = \\
&-\lambda_{disc}\ell_{disc}(G, F, D_S, D_H)  \\
&+\lambda_{cyc} \ell_{cycle}(G, F)  \\
&+\lambda_{id} \ell_{id}(G, F) 
\end{split}
\end{equation}
where $\lambda_{disc}$, $\lambda_{cyc}$ and $\lambda_{id}$ denote the weighting parameters for the discriminator, cycle  and the identity loss terms.

Here, the discriminator loss $\ell_{disc}(G, F, D_S, D_H)$ is composed of LSGAN loss \cite{mao2017least}:
\begin{equation*} \label{eq:loss_gan_x}
\begin{split}
&\ell_{disc}(G, F, D_S,D_H) =\\
& \mathbb{E}_{\yb \sim P_\Hc} \left[|D_H(\yb)\|^2_2\right]  + \mathbb{E}_{\xb \sim P_\Sc} \left[\| 1 - D_H(G_{1,0}(\xb))\|^2_2\right] \\
&+\mathbb{E}_{\xb \sim P_\Sc} \left[\|D_S(\xb)\|^2_2\right]  + \mathbb{E}_{\yb \sim P_\Hc} \left[\| 1 - D_S(G_{0,1}(\yb))\|^2_2\right]
\end{split}
\end{equation*}
where {$\|\cdot\|_2$ is the $l_2$ norm},  $G_{1,0}(\xb)$ and $G_{0,1}(\yb)$ are defined in 
\eqref{eq:GH1} and \eqref{eq:GS1}, respectively,
and $D_S$ (resp. $D_H$) is the discriminator 
that tells the fake soft kernel (resp. sharp kernel) images from real soft kernel (resp. sharp kernel) images.

The cycle loss $\ell_{cyc}(G, F)$ in \eqref{eq:loss_total} is defined as: 
\begin{equation} \label{eq:loss_cycle}
\begin{split}
\ell_{cyc}(G, F) = &\mathbb{E}_{\yb \sim P_\Hc} [\|G_{1,0}(G_{0,1}(\yb)) - \yb\|_1] \\
& + \mathbb{E}_{\xb\sim P_\Sc} [\|G_{0,1}(G_{1,0}(\xb)) - \xb\|_1]
\end{split}
\end{equation}
In addition, the identity loss in \eqref{eq:loss_total} is given by
\begin{equation} \label{eq:loss_AE}
\begin{split}
\ell_{id}(G, F) = &\mathbb{E}_{\yb \sim P_\Hc} [\|G_{1,0}(\yb) - \yb\|_1] \\
& + \mathbb{E}_{\xb \sim P_\Sc} [\|G_{0,1}(\xb) - \xb\|_1]
\end{split}
\end{equation}
Since $G_{1,0}$ is the network that converts input images to the $\Hc$ domain, $G_{1,0}(\yb)$ is in fact the
autoencoder for $\yb\in\Hc$. Similar explanation can be applied to $G_{0,1}(\xb)$ for $\xb\in \Sc$.
Accordingly, the identity  loss in \eqref{eq:loss_AE} can be interpreted as the auto-encoder loss when  the target domain images are used as inputs, and this observation is used when expanding to three domains.

\subsubsection{When three  domain training data are available}

This scenario is unique which cannot be dealt with the architecture in \cite{gu2021adain}.
Suppose that the intermediate kernel domain $\Mc$ is given by 
we assume
\begin{align*}
\Mc = 0.5\Hc+0.5\Sc
\end{align*}
implying that the $\Mc$ is in the middle of the interpolation path between $\Hc$ and $\Sc$ domain.

In this case, the training can be done by solving the following min-max optimization problem:
\begin{align}\label{eq:minmax2}
\min\limits_{G,F_e,F_d}\max\limits_{D_H,D_S,D_M}\ell_{total}(G, F_e,F_d, D_S, D_H,D_M)
\end{align}
where the total loss is given by
\begin{equation} \label{eq:loss_total2}
\begin{split}
&\ell_{total}(G, F_e,F_d, D_S, D_H, D_M) \\
=&- \lambda_{disc}\ell_{disc}(G, F_e, F_d, D_S, D_H,D_M) \\
&+\lambda_{cyc} \ell_{cycle}(G, F_e,F_d)  \\
&+\lambda_{AE} \ell_{AE}(G, F_e,F_d)  \\
&+\lambda_{sc} \ell_{sc}(G, F_e,F_d)
\end{split}
\end{equation}
Thanks to the availability of the intermediate domain $\Mc$, the discriminator loss should
be modified to include the additional domain information:
\begin{equation*} \label{eq:loss_gan_x2}
\begin{split}
&\ell_{disc}(G, F_e,F_d, D_S,D_H,D_M) =\\
& \mathbb{E}_{\yb \sim P_\Hc} \left[|D_H(\yb)\|^2_2\right]  + \mathbb{E}_{\xb \sim P_\Sc} \left[\| 1 - D_H(G_{1,0}(\xb))\|^2_2\right] \\
&+\mathbb{E}_{\xb \sim P_\Sc} \left[\|D_S(\xb)\|^2_2\right]  + \mathbb{E}_{\yb \sim P_\Hc} \left[\| 1 - D_S(G_{0,1}(\yb))\|^2_2\right]\\
&+\mathbb{E}_{\xb \sim P_\Sc} \left[\|D_S(\xb)\|^2_2\right]  + \mathbb{E}_{\zb \sim P_\Mc} \left[\| 1 - D_S(G_{0.5,1}(\zb))\|^2_2\right]\\
&+\mathbb{E}_{\zb \sim P_\Mc} \left[\|D_M(\zb)\|^2_2\right]  + \mathbb{E}_{\xb\sim P_\Sc} \left[\| 1 - D_M(G_{1,0.5}(\xb))\|^2_2\right]\\
&+\mathbb{E}_{\yb \sim P_\Hc} \left[\|D_H(\yb)\|^2_2\right]  + \mathbb{E}_{\zb \sim P_\Mc} \left[\| 1 - D_H(G_{0.5,0}(\zb))\|^2_2\right]\\
&+\mathbb{E}_{\zb \sim P_\Mc} \left[\|D_M(\zb)\|^2_2\right]  + \mathbb{E}_{\yb \sim P_\Hc} \left[\| 1 - D_M(G_{0,0.5}(\yb))\|^2_2\right]
\end{split}
\end{equation*}
where 
$G_{\alpha,\beta}$ is now defined by \eqref{eq:Gab}, and
$D_M$ is the discriminator 
that tells the fake intermediate kernel images from real  intermediate kernel images.
The cycle loss in \eqref{eq:loss_total2} should be also modified to include the intermediate domain:
\begin{equation} \label{eq:loss_cycle2}
\begin{split}
\ell_{cyc}(G, F_e,F_d) = &
\mathbb{E}_{\yb \sim P_\Hc} [\|G_{1,0}(G_{0,1}(\yb)) - \yb\|_1] \\
& + \mathbb{E}_{\xb\sim P_\Sc} [\|G_{0,1}(G_{1,0}(\xb)) - \xb\|_1] \\
&+ \mathbb{E}_{\yb \sim P_\Hc} [\|G_{0.5,0}(G_{0,0.5}(\yb)) - \yb\|_1] \\
&+ \mathbb{E}_{\xb\sim P_\Sc} [\|G_{0.5,1}(G_{1,0.5}(\xb)) - \xb\|_1] \\
&+ \mathbb{E}_{\zb \sim P_\Mc} [\|G_{0,0.5}(G_{0.5,0}(\zb)) - \zb\|_1] \\
&+ \mathbb{E}_{\zb\sim P_\Mc} [\|G_{1,0.5}(G_{0.5,1}(\zb)) - \zb\|_1] 
\end{split}
\end{equation}
The auto-encoder loss in \eqref{eq:loss_total2}, which is an extension of the identity loss in \eqref{eq:loss_AE}, is given by
\begin{equation} \label{eq:loss_AE2}
\begin{split}
\ell_{AE}(G, F_e,F_d) = &\mathbb{E}_{\yb \sim P_\Hc} [\|G_{0,0}(\yb) - \yb\|_1] \\
& + \mathbb{E}_{\xb \sim P_\Sc} [\|G_{1,1}(\xb) - \xb\|_1]\\
&+ \mathbb{E}_{\zb\sim P_\Mc} [\|G_{0.5,0.5}(\zb) - \zb\|_1] 
\end{split}
\end{equation}
Finally, one unique loss which was not present in the two domain training is the the self-consistency loss,
which imposes the constraint that the style transfer through intermediate domain should be the same as the direct conversion.
This can be represented by
 \begin{equation} \label{eq:loss_self_consistency}
\begin{split}
&\ell_{sc}(G, F_e,F_d) =\\
 &\mathbb{E}_{\yb \sim P_\Hc} [\|G_{0.5,1}(G_{0,0.5}(\yb) - G_{0,1}(\yb)\|_1] \\
& + \mathbb{E}_{\xb \sim P_\Sc} [\|G_{0.5,0}(G_{1,0.5}(\xb) - G_{1,0}(\xb)\|_1]
\end{split}
\end{equation}
These loss functions are illustrated in Fig.~\ref{fig_adain_vs_vanilla}.

\section{Methods}
\label{sec:methods}

\subsection{Data Acquisition}

{To verify the proposed continuous kernel conversion, we use the following
 dataset provided by Gachon University Gil Hospital, Korea.}

\subsubsection{Head dataset}
{Head images from 11 patients were obtained (SOMATOM Definition Edge, Siemens Healthineers, Germany). Each patient had two sets of images, each of which was generated with sharp and soft kernels, respectively. Specific details of kernels will be explained later.
Each patient data consists of around 50 slices. Accordingly, total 540 slices were achieved.  One patient was excluded because of its different image matrix size ($571\times 512$ vs. $512\times 512$).  We used seven patients for training, two for validation, the other one patient for test (44 slices). Totally, we used ten patients consisting of three men and seven women with mean age $39.7 \pm 19.1$ years.}

\subsubsection{Facial bone dataset}
{Facial bone images of 12 patients were obtained. Similar to the head dataset, each patient data was composed of reconstructed images with sharp and soft kernels for bone and brain, respectively. Each patient data involved different number of slices from 48 to 189 slices, producing 1683 slices of facial bone images for each kernel. One patient was excluded because of its different image matrix size ($512\times 534$ vs. $512\times 512$). We used eight patients for training, two for validation, the remaining one patient for test (165 slices). Totally, we used 11 patients consisting of four men and seven women with mean age $32.3 \pm 15.4$ years.} 

Additional face bone images of eight patients were obtained that include three kernel domain data. Different from the dataset above, each patient data was composed of reconstructed images with sharp, soft, and middle kernels. Each patient data includes from 167 to 209 slices. The dataset has totally 1491 slices of facial bone images for each kernel. We used five patients for training, two for validation, the other one patient for test (209 slices). Totally, the dataset was composed of five men and three women with mean age $38.6 \pm 24.5$ years.

\subsubsection{Hypopharyngeal cancer dataset}
{The dataset involved CT images of one patient suffering from hypopharyngeal cancer, which led to the demand for cartilage abnormality detection. The images covered from head to chest of its patient. The single volume was composed of 110 slices. We used a half of the slices for fine-tuning the model after training, and the other half for inference.}


\subsection{Kernels}
{For head dataset, J30s and J70h kernels were used. For facial bone dataset, J40s and J70h kernels were applied to reconstruct CT images. J30s and J40s kernels are low pass filters which are adequate for soft tissue. J70h kernel is a representative high pass filter and usually used to observe bones. 
In this paper, we refer to low pass filters such as J30s and J40s kernels as soft kernels, and high pass filters such as J70h kernel as sharp kernels.

Different from the two datasets, hypopharyngeal cancer images were reconstructed with Br44 kernel, which has similar property as a soft kernel.
In facial bone dataset for multi-domain learning, Hr40, Hr49, and Hr68 kernels were utilized to reconstruct CT images. Hr40 and Hr68 kernels correspond to soft and sharp kernels, respectively. The Hr49 kernel has intermediate properties of Hr40 and Hr68 kernels. Therefore, we refer to Hr49 kernel as the intermediate kernel in this paper.}

\subsection{Network Architecture}

\subsubsection{Autoencoder}
The autoencoder  used in this paper is based on the U-Net architecture with pooling layer implemented
by polyphase decomposition \cite{kim2019mumford}.   It has skip connections between encoder and decoder parts, which enable inputs and outputs to share information. Also, the input is added pixel-wise to the output at the end of the network. The network, therefore, learns residuals.

\begin{figure}[!hbt]
	\centerline{\includegraphics[width=0.9\columnwidth]{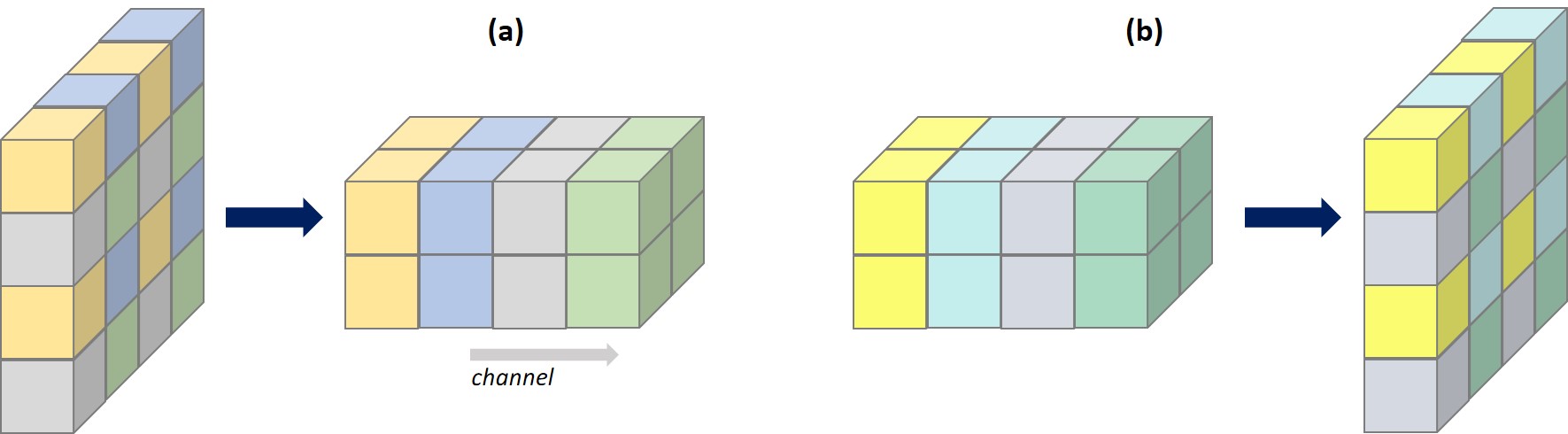}}
	\vspace*{-0.2cm}
	\caption{(a) Pooling layer using polyphase decomposition process. Pixels are divided into four groups and each group composes one image of a reduced size. Then, the groups are stacked along channel direction. (b) Unpooling layer using polyphase recomposition. }
	\label{fig_subpixel_pooling}
\end{figure}

In conventional encoder-decoder networks, the input images go through several down-sampling layers until a bottleneck layer, in order for a network to extract low frequency information. 
Through the down-sampling layers, the network necessarily loses significant amount of information which plays a crucial role in auto-encoder learning. Therefore, instead of using pooling layers, we used all the given information which can be achieved by lossless decomposition using polyphase decomposition as shown in Fig.~\ref{fig_subpixel_pooling}(a).
Specifically, at the layers where a pooling operation is required, we arranged all the pixels into four groups. 
It can be thought as having a 2$\times$2 filter with stride of 2 in order not to make overlapping. The first pixels of each sub-region gather together to form new output.  Also, the second pixels assemble another output. In the same manner, the third and fourth pixels make the third and fourth outputs, respectively. These four outputs are stacked along a channel direction.  After this sub-pixel pooling operation, the size of final output would be reduced by half while the number of channels would increase fourfold.  

\begin{figure}[!b]
	\centerline{\includegraphics[width=0.7\columnwidth,height=3cm]{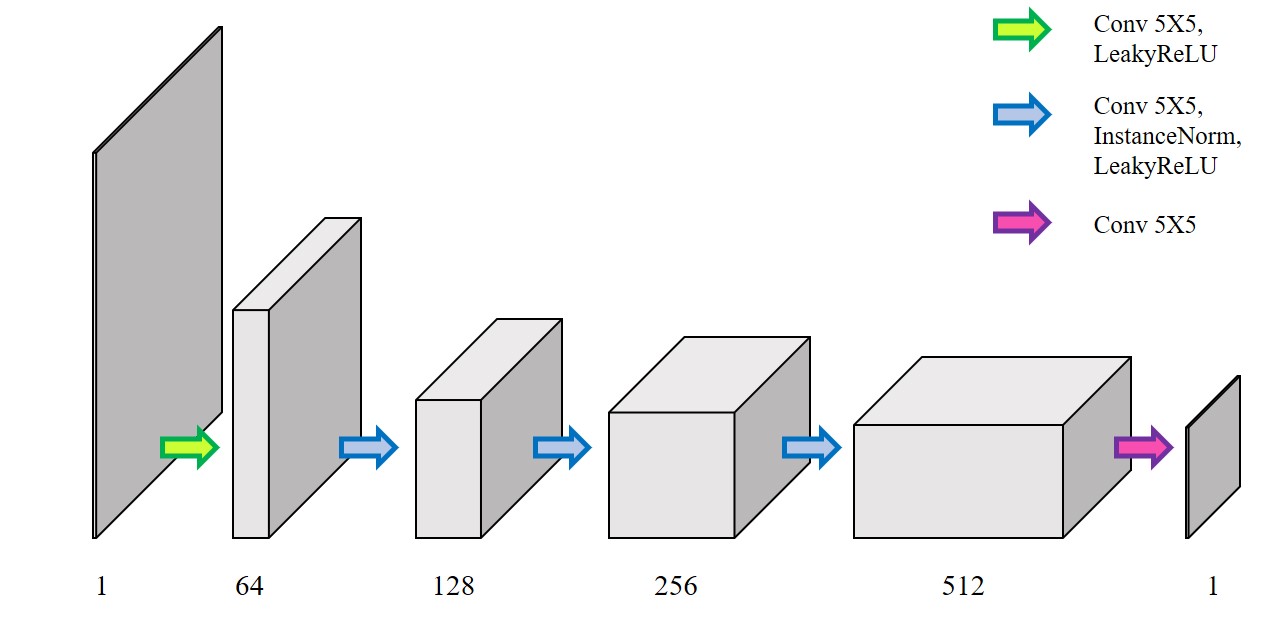}}
	\caption{Architecture of the discriminator.}
	\label{fig_discriminator}
\end{figure}

Unpooling operation using polyphase recomposition can be done in exactly the opposite way of  polyphase pooling operation as shown in Figure \ref{fig_subpixel_pooling}(b).
Interestingly, unpooling operation in a decoder part can be extended to a transposed convolution. Transposed convolution involves both polyphase recomposition  and filtering operation. While polyphase unpooling requires fixed position for each pixel, transposed convolution does not have any positional condition to follow. When using transposed convolution, the network parameters are, therefore, chosen to be optimal without any specific restriction. This results in experimentally better performance of the network with transposed convolution compared to unpooling operation using polyphase recomposition. Thus, we used transposed convolution  in our proposed Polyphase U-Net.

\subsubsection{Discriminator}

As a discriminator, we used PatchGAN \cite{isola2017image} which shows good performance at capturing high frequency information because it focuses only on the scale of its patches, not the entire image.
The illustration of the discriminator is shown in Figure \ref{fig_discriminator}. The input whose size is $128 \times 128$ with one channel passes through a convolution layer with stride two. Then, next convolution layer with stride two gets the feature map of size $62 \times 62$. The feature maps go through two successive convolution layers with stride one. Finally, the output is convolved with the last convolution layer. Kernel size of all the convolution layers in the discriminator is $5 \times 5$. Final output size is $24 \times 24$ which was chosen empirically.

\subsubsection{AdaIN Code Generator}

Details about the architecture of AdaIN code generator are illustrated in Figs. \ref{fig_2domain_generator_architecture} and \ref{fig_generator_architecture}. In the two-domain switchable cycleGAN (Fig. \ref{fig_2domain_generator_architecture}), only one shared code generator is implemented and connected to both encoder and decoder parts of the generator. One vector of size $128$ is input to the shared code generator. The shared code generator includes four fully connected layers with output size of 64. The final output code is given to 10 convolution blocks in the generator of cycleGAN. Since these convolution blocks have different number of channels, mean and variance code vectors are generated separately for each convolution block. For the mean code vectors, one fully connected layer is applied. ReLU activation layer is applied in addition to one fully connected layer for the variance code vector because of non-negative property of variance.  Meanwhile, in case of three domain conversion, we utilized switchable cycleGAN with split AdaIN, which requires source domain and target domain code generators for encoder and decoder parts of the U-net, respectively, as shown in Figure \ref{fig_generator_architecture}. 
All the other details are same as in two-domain switchable cycleGAN.

\subsection{Training detail}
The input images were randomly cropped into small patches of size $128\times 128$ during the training. They were also randomly flipped both horizontally and vertically. Despite small size of dataset, patch-based training with random flipping provided  data augmentation and enabled more stable training \cite{oh2020deep}. The learning rate was set as $10^{-4}$ and $10^{-5}$ for Head and Facial bone dataset, respectively. The Adam optimization algorithm \cite{kingma2014adam} was used, and the momentum parameters $\beta_1 = 0.9, \beta_2 = 0.999$. We saved models with the best quantitative results with validation data set and tested the models using 5-fold cross validation scheme. In case of facial bone dataset for multi-domain learning, 3-fold cross validation was conducted. We implemented the networks using PyTorch library \cite{paszke2017automatic}. We trained the networks on two NVIDIA GeForce GTX 2080 Ti.

\subsection{Comparative studies}
{First, we compared our algorithm with classical kernel conversion approaches.  Classical methods considered kernel conversion as kernel smoothing and sharpening. Specifically,  to convert sharp kernel images into soft kernel ones,  frequency band decomposition and alteration of energy in each band were performed \cite{gallardo2016normalizing}. For the conversion from soft to sharp} kernel, we sharpened the soft kernel images using Laplacian filter kernel and Wiener-Hunt deconvolution method with the estimated point spread function \cite{al2012reducing}.

{Moreover}, we compared  our proposed model with supervised learning method. For supervised learning method, the same generator
architecture as our method was used \cite{kim2019mumford} and  the mean squared error loss between output and ground truth images was used. Two different networks were trained separately for two opposite directional kernel translation. Batch size was 8 and 32 for Head and Facial bone dataset, respectively. Other settings for training were same as the proposed model. 

In addition to the supervised learning method, we conducted comparative studies using vanilla cycleGAN \cite{zhu2017unpaired}. Again,
same U-Net architecture with polyphase decomposion was used for two generators.  Due to the need to train two distinct generators, the total number of trainable weights increased compared to our method (see Figure \ref{fig_adain_vs_vanilla}). Batch size was 8 and 40 for Head and Facial bone dataset, respectively. The other training details were  same as the proposed model.

\subsection{Evaluation metrics}
Since ground-truth data are available for the two types of dataset, we used the peak signal to noise ratio (PSNR) as our quantitative metric, which is defined as follows:
\begin{equation}
PSNR = 10\log_{10}\left(\frac{MAX_{x}^2}{MSE}\right)
\end{equation}
\begin{equation}
MSE = \frac{1}{N_{1}N_{2}}\sum_{i=0}^{N_{1}-1} \sum_{j=0}^{N_{2}-1} [x_{i,j}-\hat{x}_{i,j}]
\end{equation}
where $N_{1}$ and $N_{2}$ are row and column dimensions of the images, $x_{i,j}$ denotes the $(i,j)$-th pixel,
 and $MAX_{x}$ is the maximum possible pixel value of image $x$.  We also used structural similarity (SSIM) index \cite{wang2004image} which is defined as
\begin{equation}
SSIM(x, \hat{x}) = \frac{(2m_{x}m_{\hat{x}}+c_1)(2\sigma_{x\hat{x}}+c_2)}{(m_{x}^2+m_{\hat{x}}^2 + c_1)(\sigma_{x}^2+ \sigma_{\hat{x}}^2+ c_2)}. 
\end{equation}
where $m$ is average of the image, $\sigma$ is variance of the image, and $\sigma_{x\hat{x}}$ is covariance of the images $x$ and $\hat{x}$. The two variables $c_{1} = (k_{1}L)^2$ and $c_{2} = (k_{1}L)^2$ are used to stabilize the division where $L$ is the dynamic range of the pixel intensities and $k_{1} = 0.01, k_{2} = 0.03$ by default.

\begin{table}[hbt!]
\centering
\caption{Quantitative comparison of various methods in two-domain learning}
\setlength{\tabcolsep}{4pt}
\begin{tabular}{p{110pt} c c c c}
	\hline
	\multirow{2}{*}{Head}&  \multicolumn{2}{c}{PSNR}& \multicolumn{2}{c}{SSIM} \\
	& sharp & soft & sharp & soft \\
	\hline
	\hline
	Classical method & 12.4335 & 14.4620 & 0.6633 & 0.7401 \\
	Supervised (MSE Loss) & 32.6624 & 24.2981 & 0.9022 & 0.8336 \\
	Vanilla CycleGAN & 30.8758 & 21.9194 & 0.8913 & 0.8161 \\
	Switchable CycleGAN & 31.6671 & 23.9193 & 0.8922 & 0.8766 \\
	\hline
	\multicolumn{5}{c}{} \\
	\hline
	\multirow{2}{*}{Facial Bone}&  \multicolumn{2}{c}{PSNR}& \multicolumn{2}{c}{SSIM}
	\\
	& sharp & soft & sharp & soft \\
	\hline
	\hline
	Classical method & 13.1227 & 10.4685 & 0.5198 & 0.5369 \\
	Supervised (MSE Loss) & 28.3261 & 21.1531 & 0.8079 & 0.8088 \\
	Vanilla CycleGAN & 25.5337 & 17.4077 & 0.7135 & 0.7966 \\
	Switchable CycleGAN & 26.6328 & 19.5712 & 0.7563 & 0.8336 \\
	\hline
\end{tabular}
\label{tab_result_main}
\end{table}

\begin{figure*}[!hbt]
	\centerline{\includegraphics[width = 1.9\columnwidth]{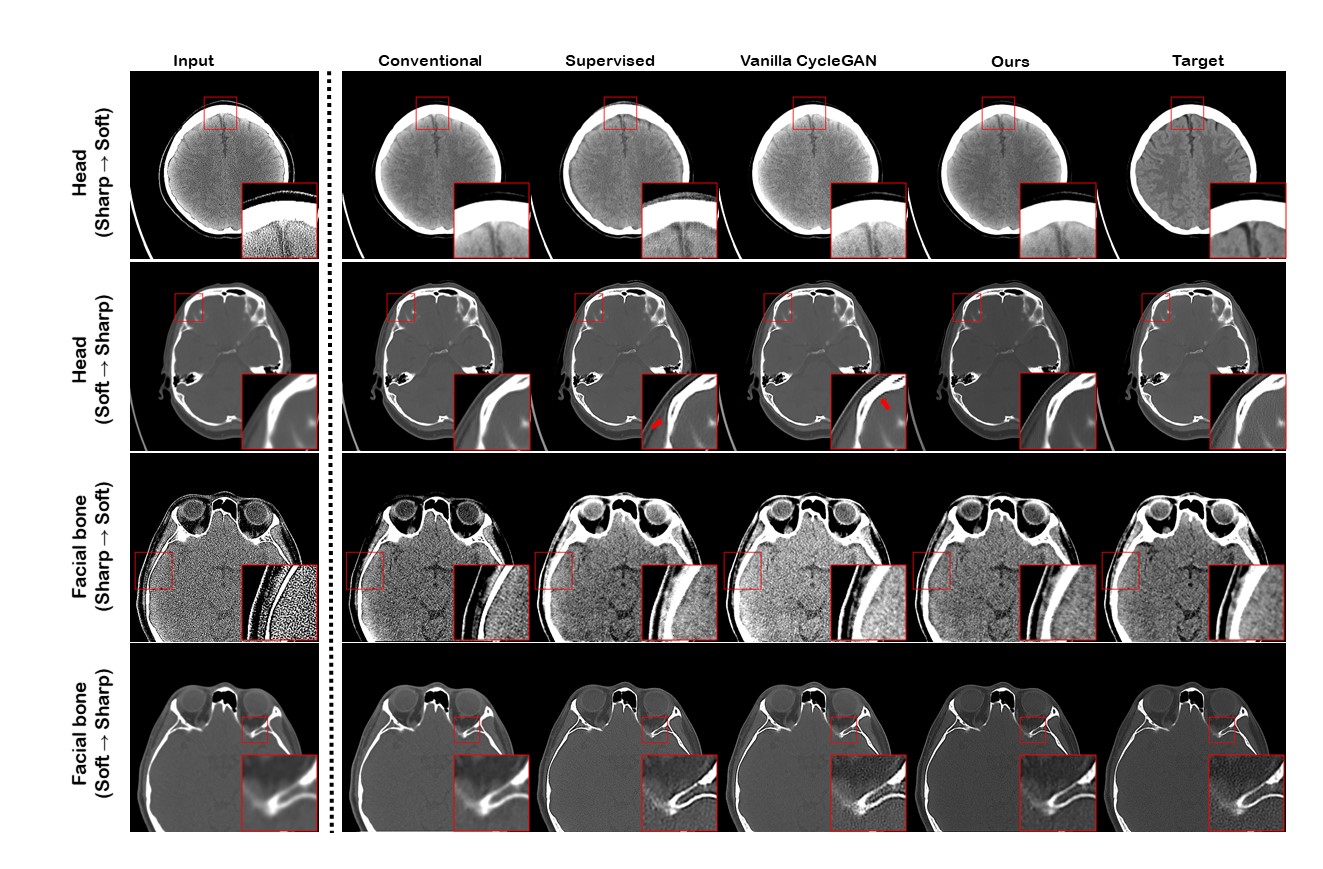}}
	\vspace*{-0.5cm}
	\caption{Kernel conversion results in two-domain by the classical method, supervised learning, vanilla cycleGAN, and  the proposed switchable cycleGAN. The first two rows are from Head dataset and the last two rows are from Facial bone dataset.}
	\label{fig_result_main}
\end{figure*}

\section{Experimental results}
\label{sec:results}


\subsection{Comparison with previous methods}
First, we compared our proposed model with classical kernel conversion method, supervised learning method, and vanilla cycleGAN \cite{zhu2017unpaired}. 
{In Figure~\ref{fig_result_main} and Table~\ref{tab_result_main}, the images generated from the classical method showed the worst performance in both quantitative and qualitative perspectives.}
Although higher PSNR and SSIM values were obtained for supervised learning in Table \ref{tab_result_main},  visual investigation shows that some blurring artifacts are presented with the supervised learning as shown in Figure \ref{fig_result_main}.
 More details are as follows:

\begin{figure*}[!t]
	\centerline{\includegraphics[width = 2\columnwidth]{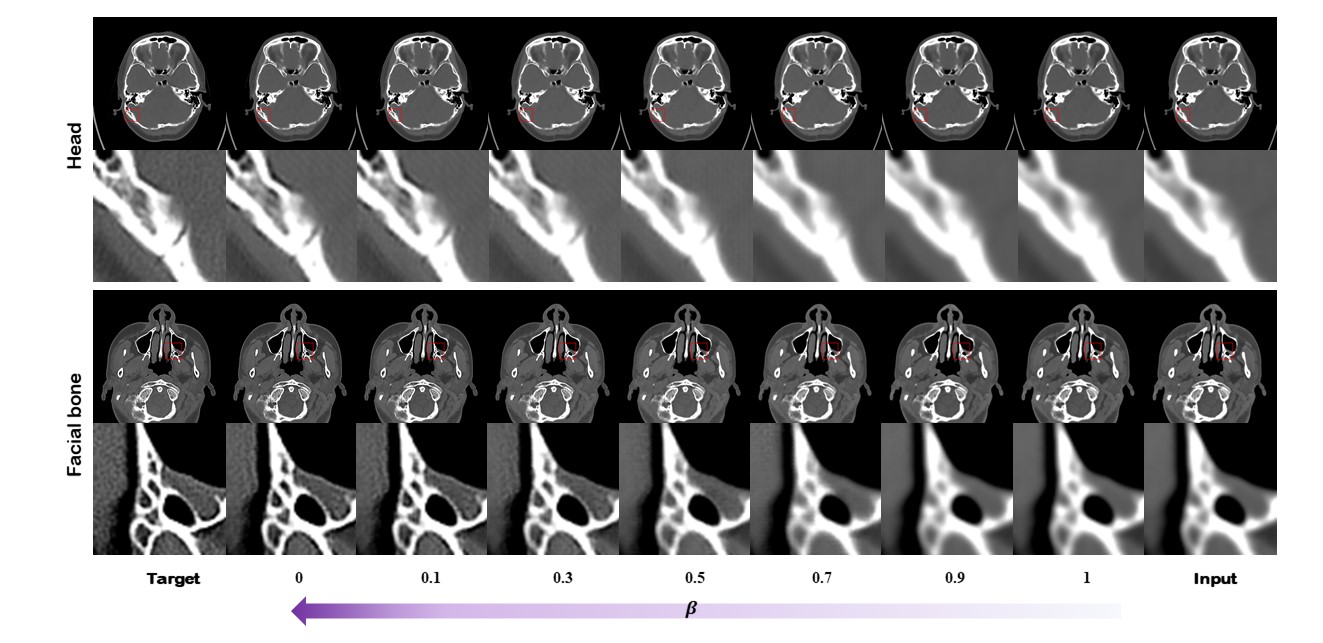}}
	\caption{Continuous interpolation of kernel images along an optimal transport path.}
	\label{fig_result_alpha_fakeH}
\end{figure*}

\begin{figure*}[!t]
	\centerline{\includegraphics[width = 2\columnwidth]{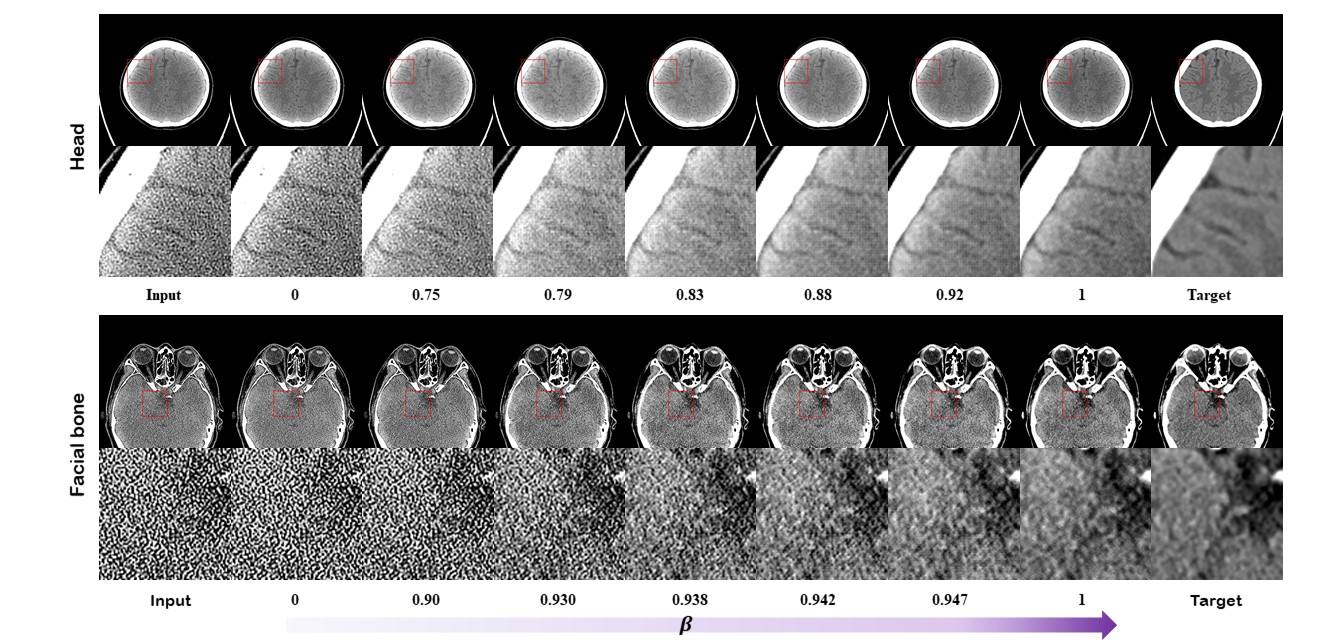}}
	\caption{Continuous interpolation of kernel images along an optimal transport path.}
	\label{fig_result_alpha_fakeS}
\end{figure*}

\subsubsection{Soft kernel from sharp kernel}
Image translation from sharp kernel images to soft kernel images is shown in the first and third rows of Figure \ref{fig_result_main}. Soft kernel images should clearly identify soft tissues such as blood vessels, not bones. It is shown that performance of  our proposed method is obviously better than those of supervised and vanilla cycleGAN methods. In particular, as can be seen in the first row of Fig. \ref{fig_result_main}, the grid-like artifacts appeared on the result of the supervised model. The images generated from cycleGAN could not follow the data distribution and the pixel intensities around the bone were significantly increased. However, these artifacts were not shown in the results of our method.
We believe that the artifacts for the supervised learning and vanilla cycle GAN may occur because of the limited training data set.
Nonetheless,  we have found that our proposed model has its strength in capturing the data distribution much more effectively.

{The third row of Fig. \ref{fig_result_main} also shows that the bone shapes reconstructed by the conventional method are obviously different from the target images.}
In addition, the synthesized soft kernel images by the traditional method and the supervised learning model show significantly more noise compared to the other unsupervised learning methods. 
The results clearly show that our model was better at translating sharp kernel images into soft kernel images than the supervised or vanilla cycleGAN.

\subsubsection{Sharp kernel from soft kernel}
Sharp kernel images are usually used to gather information about clear bone outlines. The images generated from soft kernel images should contain information about a clear delineation of the bone. 

The results of the image translation from soft kernel images to sharp kernel images are shown in the second and fourth rows of Figure \ref{fig_result_main}. {The traditional method sharpened the input soft kernel images, but their sharpness was far below that of the target images. In the meantime, the results of all three deep-learning} methods were similar. 
Not only could they enhance  sharp bones outlines from blurry, soft kernel images, but they could also follow the texture of sharp kernel domain.
However, some artifacts occurred along bones in the supervised and cycleGAN methods (Figure. \ref{fig_result_main}).

\begin{figure*}[!t]
	\centerline{\includegraphics[width = 2\columnwidth]{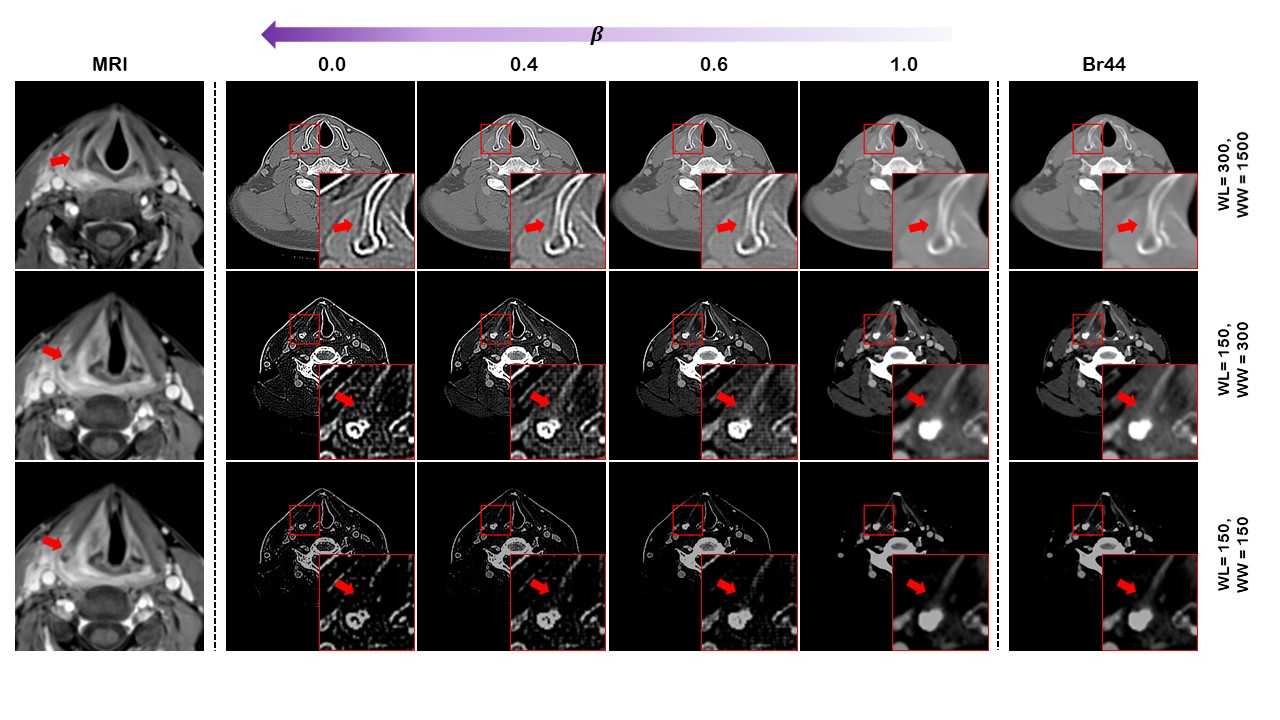}}
	\caption{Cartilage abnormality is clearly seen as $\beta$ value varies. The left most column is MR images with contrast enhancement. Right thyroid cartilage tumor infiltration can be observed from MR images whereas it is hard to detect on CT images which are on the right most column. They were reconstructed with Br44 kernel which has properties of the soft kernel. The images in the middle were generated from the images in the right most column. Compared to MR images, the generated CT images showed a comparable level of diagnostic ability to check cartilage abnormality. With input kernel images on the right most column, sharp kernel images were generated using $\beta=0$, and soft kernel images using $\beta=1$}. 
	\label{fig_result_alpha_cartilage}
\end{figure*}

\begin{figure*}[!hbt]
	\centerline{\includegraphics[width = 2\columnwidth]{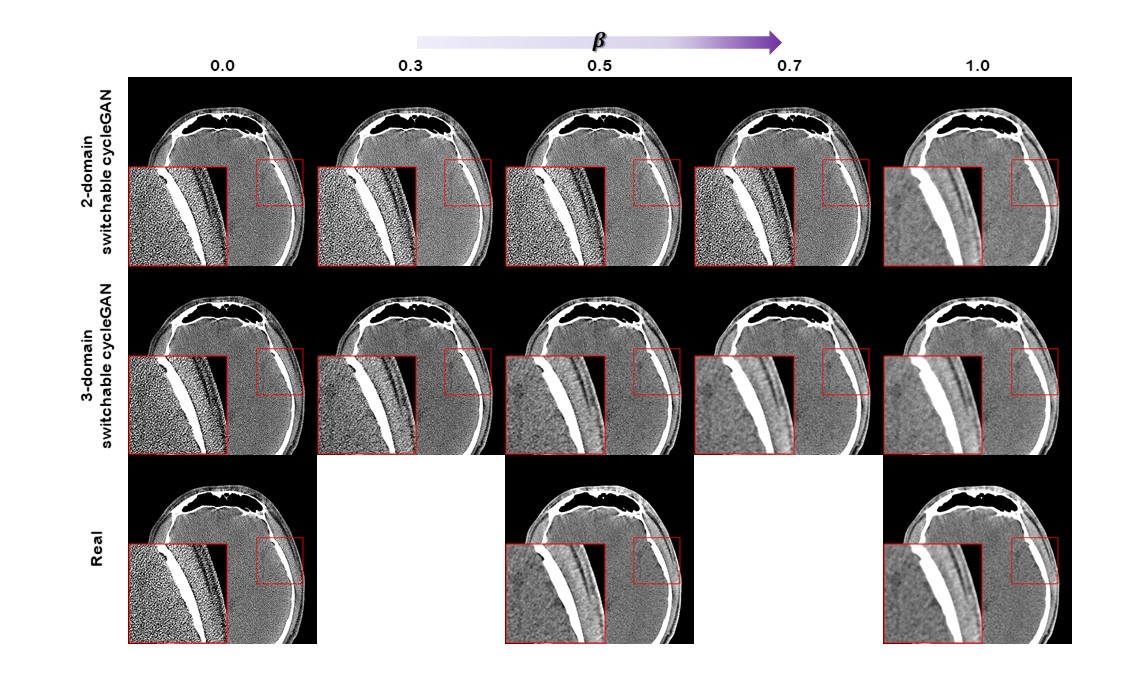}}
	\caption{Interpolation from sharp to soft kernel results in two-domain and three-domain learning. 
	The first row is the switchable cycleGAN with single AdaIN code generator while the second row is the switchable cycleGAN with split AdaIN. The last row is real CT images. The right and middle ones correspond to target images and the left one is the input image.}
	\label{fig_result_3domain_soft}
\end{figure*}

\subsubsection{Contributing factors of AdaIN to improved quality}

Image translation by switchable cycleGAN from soft kernel to sharp kernel images is  actually same as vanilla cycleGAN since it utilizes mean and variance of its own instead of those estimated from AdaIN. Thus, the sharp kernel images 
 that were generated with our model showed in qualitative and quantitative results an almost equivalent quality to that with the vanilla cycle GAN. 
 For SSIM results on the head data, paired $t$-test was performed on the sharp images generated by vanilla and switchable cycleGAN.
 The $p$-value was larger than $0.05$, which means that our switchable cycleGAN generated sharp kernel images that were not significantly different from
 those by  vanilla cycleGAN in terms of SSIM metric (Table \ref{tab_result_main}). 

On the other hand, the model in the other direction, which translates sharp kernel images into soft kernel images, contains AdaIN, so that the mean value and variance of the sharp kernel domain are scaled or added. It should be noted that the image quality of our proposed model has been significantly improved in relation to that of vanilla CycleGAN. It is supported in both qualitative and quantitative results.

{In particular, artifacts could be observed in soft kernel images generated from the vanilla cycleGAN method, and this made it difficult to accurately distinguish the boundaries between bone and soft tissue. In addition, the pixel intensities around the bone have been increased in the vanilla cycleGAN images. However, these artifacts disappeared in the generated soft kernel images of our model (Figure \ref{fig_result_main}).} 
In addition, the evaluation metrics when generating soft kernel images have been significantly improved. 
For $t$-test on the PSNR and SSIM results between the vanilla cycleGAN and the proposed method, the $p$-values were less than 0.01 when generating soft kernel images from the head dataset (Table \ref{tab_result_main}).
Furthermore, when comparing PSNR results from the soft kernel images of the facial bone dataset, the $p$-value was less than $0.05$. 
From these analyses, we could see that switchable cycleGAN showed significantly improved performance in generating soft kernel images compared to the vanilla cycleGAN. The effectiveness of AdaIN was clearly demonstrated by this different increase in image quality between two opposite directions.

Although the small data set can be a major factor in overfitting the network, thanks to AdaIN the number of model parameters in our model has been halved and model learning has been proven to be more stable \cite{gu2021adain}.

\subsection{Interpolation between two kernel images}
During the training phase, we set $\beta$ as one if soft kernel images are generated from sharp kernel images, and $\beta$ as zero if sharp kernel images are generated from soft kernel images. 
Once the training was done, at the inference phase different {$\beta$} values can be applied to generate
interpolating kernel images along an optimal transport path 
(Fig. \ref{fig_result_alpha_fakeH} and \ref{fig_result_alpha_fakeS}). 
Interestingly, we were able to see objects that could not be observed with previous kernels.

\begin{figure*}[!hbt]
	\centerline{\includegraphics[width = 2\columnwidth]{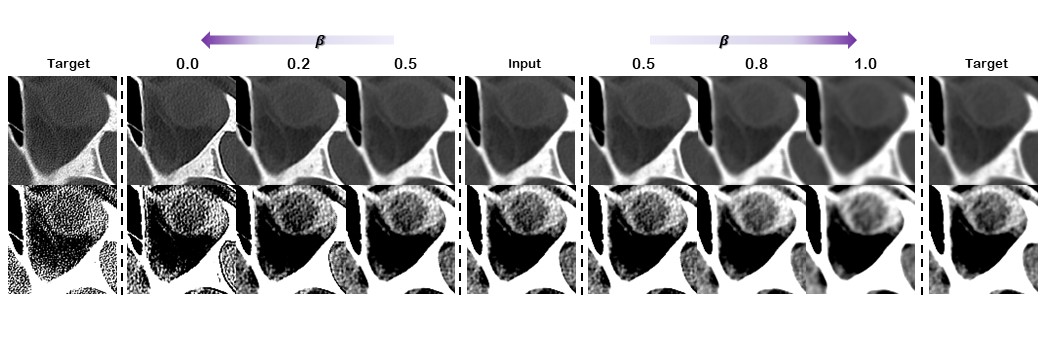}}
	\caption{Interpolation from middle to sharp or soft kernel images in three-domain learning. The window level and width of the first row is 400 and 1500 and those of the second row are 50 and 120 (HU).}
	\label{fig_result_3domain_middle}
\end{figure*}

{In particular, once our proposed method is applied, 
cartilage abnormality can be detected as shown in Figure \ref{fig_result_alpha_cartilage}. For hypopharyngeal cancer case, it is important to check whether thyroid cartilage is invaded or not.  This identification usually relies on magnetic resonance imaging since it is hard to observe the abnormality from CT images. Since the cartilage has high attenuation like bone, sharp kernel images could be potentially useful in determining if the cancer has penetrated or not.  Unfortunately, CT images are not reconstructed with sharp kernels to evaluate this type of cancer. Instead, soft-kernel CT images are usually collected for other
diagnostic purpose.} With the help of our proposed method, we conjectured that discontinuity in outer border of cartilage could be discovered from the generated images from soft kernel images.

Specifically, the hypopharyngeal cancer data contains only Br44 kernel images which have soft kernel properties, which means there exists no target domain images. Therefore, we trained the model with some slices of these Br44 kernel images and J70h kernel images from Facial bone dataset as target domain data.
Specifically, our model is trained to  generate J70h kernel images when $\beta=0$, and Br44 kernel images when $\beta=1$. 

In Figure \ref{fig_result_alpha_cartilage},
for $\beta=0$, overshoots and undershoots occurred around the bone outlines, which hindered close observation on bone shapes. However, more accurate examination could become possible by adjusting $\beta$ values to be larger than zero. Specifically, the images from $\beta$ larger than zero could be generated with smoother texture compared to those from $\beta=0$. This resulted in clear delineation with no more undershoot or overshoot around the bone but maintaining its shape as shown Figure \ref{fig_result_alpha_cartilage}.
This example clearly shows the clinical usage of the continuous conversion of the filter kernels along the optimal transport path.
It is reminded that the simple image domain interpolation does not provide such synergistic information as ours since it is a simple mixing of image domain textures. See Section~\ref{sec:discussion} for more details.

\begin{figure*}[!hbt]
	\centerline{\includegraphics[width = 2\columnwidth]{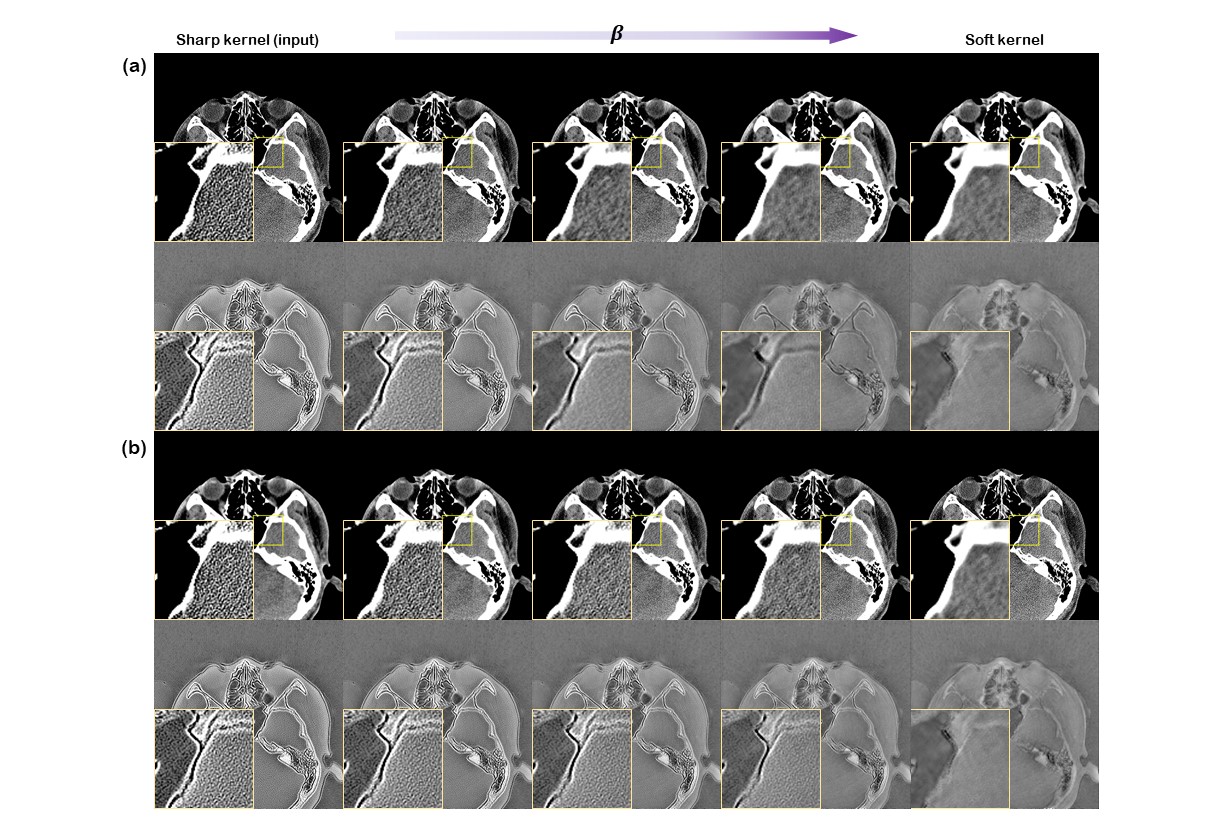}}
	\caption{(a) AdaIN-based feature domain interpolation. (b) Image domain interpolation. The experiment was conducted on facial bone dataset for multi-domain learning. The left-most column is sharp kernel images used as input and the other columns are the generated images. The second and the fourth rows correspond to difference images between the generated images and real soft kernel images. The difference image window is (-200, 200) HU.} 
	\label{fig_interpolation}
\end{figure*}

\subsection{Three domain learning}
One of the limitations of switchable CycleGAN with two domains is that the effect of $\beta$ is not gradual. Although the generated soft kernel images are subject to gradual changes in the range from 0.9 to 1.0 in Fig.~\ref{fig_result_alpha_fakeS}, outside of this range, the generated images showed no noticeable changes.

Therefore, another important contribution of this paper is to show that this non-uniform dependency on $\beta$ can be addressed if the intermediate domain data is available.
Specifically, by utilizing the  middle kernel images during the training, the generated images with $\beta = 0.5$ could have similar property of that of real middle kernel images. This can be confirmed in Figure \ref{fig_result_3domain_soft}. Here,
the images at $\beta = 0.5$ from  three-domain switchable cycleGAN were similar to real images of Hr49 kernel,  whereas the two domain switchable cycleGAN produces no noticeable changes between 0$\sim$ 0.7. 
In fact, the generated images changed gradually from $\beta = 0$ to $\beta = 1$ in the three-domain learning.

In addition, generating sharp or soft kernel images from middle ones is only possible by using three domain data (Figure \ref{fig_result_3domain_middle}). Here, intermediate kernel images can be utilized to generate other kernel images, which further helped to enhance the performance of the network with self-consistency loss. Specifically, the middle kernel images generated from sharp kernel images can be used as input to generate soft kernel images. Then these soft kernel images are forced to be identical to the soft kernel images generated directly from sharp kernel ones. This constraint helped to improve the three domain conversion of the proposed model. The role of the self-consistency loss is described in more detail later.

Additionally, the performance of the model slightly enhanced with the three domain learning,  as can be seen in Table \ref{tab_result_3domain}
in terms of PSNR and SSIM values. 
Accordingly, we can conclude that  three domain conversion would be preferable with respect to the quality of the generated images and the effective range of interpolation.

\begingroup
\renewcommand{\arraystretch}{1.3} 
\begin{table}[hbt!]
\centering
\caption{Quantitative comparison of various methods with facial bone dataset for multi-domain learning}
\setlength{\tabcolsep}{4pt}
\begin{tabular}{c p{85pt} c c c c}
	\hline
	\multirow{2}{*}{Domain} &
	\multirow{2}{*}{Method} &  
	\multicolumn{2}{c}{PSNR} & 
	\multicolumn{2}{c}{SSIM} \\
	& & sharp & soft & sharp & soft \\
	\hline
	\hline
	\multirow{5}{*}{3} & 
	Switchable cycleGAN with self-consistency loss (Ours) & 
	\multirow{2}{*}{\textbf{25.7611}} & 
	\multirow{2}{*}{\textbf{19.2320}} &
	\multirow{2}{*}{0.7137} &
	\multirow{2}{*}{\textbf{0.8524}} \\ 
	 & 
	Switchable cycleGAN without self-consistency loss &
	\multirow{2}{*}{24.2031} & 
	\multirow{2}{*}{16.6467} &
	\multirow{2}{*}{0.6658} &
	\multirow{2}{*}{0.8194} \\ 
	\hline
	\multirow{2}{*}{2} & Switchable cycleGAN & 25.3679 & 
	18.8973 & \textbf{0.7345} & 0.8139 \\ 
	& Vanilla cycleGAN & 25.2707 & 17.6606 & 0.7084 & 0.8240 \\
	\hline \\
\end{tabular}
\label{tab_result_3domain}
\end{table}
\endgroup

\section{Discussion}
\label{sec:discussion}

\subsection{Strengths of Switchable cycleGAN}

\subsubsection{Difference from image domain interpolation}
One of the strengths of switchable cycleGAN is that it can generate images  between two image domains through feature domain interpolation. Specifically, the feature domain interpolation between two domains can be performed using AdaIN code vector interpolation. 
Figure \ref{fig_interpolation} illustates the difference between the feature domain and image domain interpolation. When generating soft kernel images from sharp kernel ones, we could not get middle kernel images from image domain interpolation. The noise level was not reduced enough to get smooth texture of middle kernel. Also, these noises were spread on the entire area until $\beta$ values became one to generate soft kernel images. On the other hand, as can be seen from the difference images in Figure \ref{fig_interpolation},
our feature domain interpolation gradually changes the texture of  the images.

\subsubsection{Robustness to dataset size}
\begin{figure}[!t]
	\centerline{\includegraphics[width = \columnwidth]{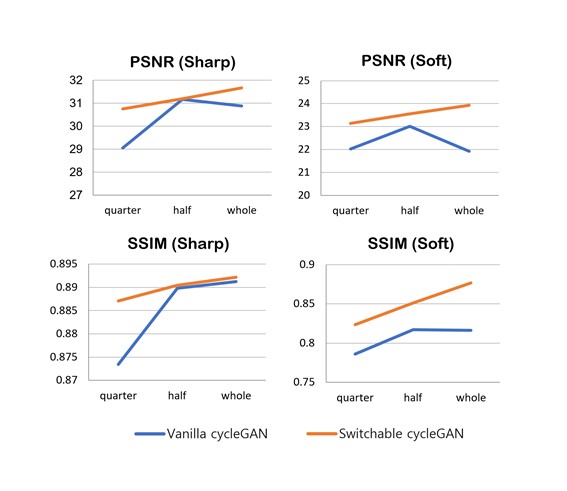}}
	\caption{Comparison between vanilla cycleGAN and switchable cycleGAN according to dataset size. The two methods have similar number of trainable parameters. }
	\label{fig_dataset}
\end{figure}

Since switchable cycleGAN shares a single generator, the number of parameters of the model is reduced, which results in robust training even on small dataset. Similar to the experiments conducted in \cite{gu2021adain},
we also investigate the robustness to the training dataset size using Head dataset. 
In  Figure \ref{fig_dataset}, vanilla cycleGAN showed sudden decreases in both PSNR and SSIM  performance
as dataset size decreased. 
However, the switchable cycle GAN showed a moderate drop in performance with the reduced training data set.
In particular, our method trained with only a quarter of the data set leads to better performance compared to the vanilla cycleGAN trained with the entire data set.

\begin{table}[hbt!]
\centering
\caption{Quantitative comparison on the effect of the number of parameters of the generator}
\setlength{\tabcolsep}{4pt}
\begin{tabular}{p{120pt} c c c c}
	\hline
	\multirow{2}{*}{Head}&  \multicolumn{2}{c}{PSNR}& \multicolumn{2}{c}{SSIM} \\
	& sharp & soft & sharp & soft \\
	\hline
	\hline
	Vanilla CycleGAN (half parameters) & 31.3088 & 23.3729 & 0.8909 & 0.8461 \\
	Switchable CycleGAN & \textbf{31.6671} & \textbf{23.9193} & \textbf{0.8922} & \textbf{0.8766} \\
	\hline
\end{tabular}
\label{tab_half_param}
\end{table}

\begin{figure*}[!t]
	\centerline{\includegraphics[width = 1.8\columnwidth]{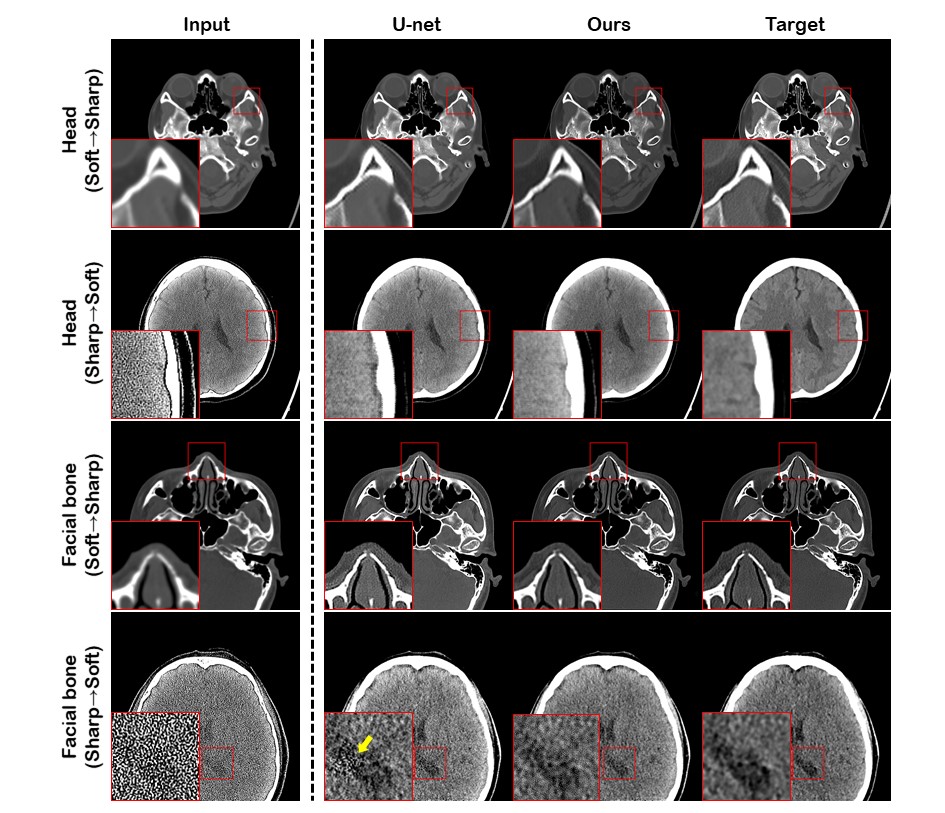}}
	\caption{Comparison between U-Net \cite{ronneberger2015u} and Polyphase U-Net. }
	\label{fig_result_Polyphase}
\end{figure*}

\subsubsection{The number of parameters of the generator}
We could expect the performance improvement thanks to the shared generator of the the switchable cycleGAN. 
This improvement results from the smaller number of parameters in the network as well as from the shared generator.
Though the network size reduction provided robustness to the training dataset size, we found that the benefit from
the shared network architecture is more profound from following experiments.

Specifically, as can be seen from the Table \ref{tab_half_param}, switchable cycleGAN showed significantly increased performance over the vanilla cycleGAN. From the paired $t$-test between the vanilla cycleGAN and the proposed method, $p$-value was less than $0.05$ on both PSNR and SSIM results of the soft kernel images. Also, when comparing PSNR results on the sharp kernel images between the vanilla cycleGAN and the proposed method, $p$-value was less than $0.05$.
In this comparative experiment, the vanilla cycleGAN involves two different generators with the number of parameters reduced by half
so that the total number of trainable parameters are about the same as our switchable cycleGAN.
Nonetheless, our switchable cycleGAN outperformed vanilla cycleGAN. This suggests that shared generator is a key that has contributed to learning the common representation of data and improving the performance of the model.

\subsection{Effect of Polyphase U-Net}

{Here, we investigated  the differences between U-Net and Polyphase U-Net as a generator in our switchable cycleGAN.
For a fair comparison with the original U-Net \cite{ronneberger2015u}, we reduced the number of channels so that
the total number of parameters in the Polyphase U-Net is around 30 millions.
 As shown in Figure \ref{fig_result_Polyphase}, Polyphase U-Net showed enhanced performance compared to the U-Net \cite{ronneberger2015u}. 
The images generated from the Polyphase U-Net were very similar to the target images with a sharp kernel in terms of outlines of bones, but also textures.

Specifically, the shapes of the bones followed the target images well and were clear enough to identify the small structures. In addition, it was difficult to identify a broken nasal bone with the image reconstructed with a soft kernel
(Figure \ref{fig_result_Polyphase}, third row). Only the existence of the bone could  be confirmed. 
To our expectation, the shape of the broken bone is clearly shown  in the kernel conversion by our method using the Polyphase U-Net architecture. However, 
the first and the third row of Figure \ref{fig_result_Polyphase} show that some irrelevant patterns around bones appeared when the conventional U-Net \cite{ronneberger2015u} is used as a generator. 
} 
In addition, in the generation of soft kernel images, 
Polyphase U-Net also provides slightly better results than the conventional U-Net. As shown in the fourth row of Figure \ref{fig_result_Polyphase}, the texture of the images generated by U-Net, which should be as smooth as the target image, is noisy. 

The improved performance of the Polyphase U-Net was also reflected in the quantitative results (Table \ref{tab_result_Polyphase}). Since the problem of information loss during pooling operation was resolved in the Polyphase U-Net, it showed better performance compared to the original U-Net.

\begin{table}[hbt!]
\centering
\caption{Quantitative Comparison between U-Net and PolyPhase U-Net}
\setlength{\tabcolsep}{4pt}
\begin{tabular}{p{110pt}| c c c c}
	\hline
	\multirow{2}{*}{Head}&  \multicolumn{2}{c}{PSNR}& \multicolumn{2}{c}{SSIM} \\ 
	& sharp & soft & sharp & soft \\
	\hline
	\hline
	CycleGAN with U-Net & 31.2706 & 23.2257 & 0.8887 & 0.8615 \\
	CycleGAN with PolyPhase U-Net & \textbf{31.6671} & \textbf{23.9193} & 				\textbf{0.8922} & \textbf{0.8766} \\
	\hline
	\multicolumn{5}{c}{} \\
	\hline
	\multirow{2}{*}{Facial Bone}&  \multicolumn{2}{c}{PSNR}& \multicolumn{2}{c}{SSIM}
	\\
	& sharp & soft & sharp & soft \\
	\hline
	\hline
	CycleGAN with U-Net & 26.1938 & 18.7120 & 0.7490 & 0.7912 \\
	CycleGAN with PolyPhase U-Net & \textbf{26.6328} & \textbf{19.5712} & 
	\textbf{0.7563} & \textbf{0.8336} \\
	\hline
\end{tabular}
\label{tab_result_Polyphase}
\end{table}

\subsection{Role of self-consistency loss in multi-domain learning}
\begin{figure*}[!t]
	\centerline{\includegraphics[width = 2.1\columnwidth]{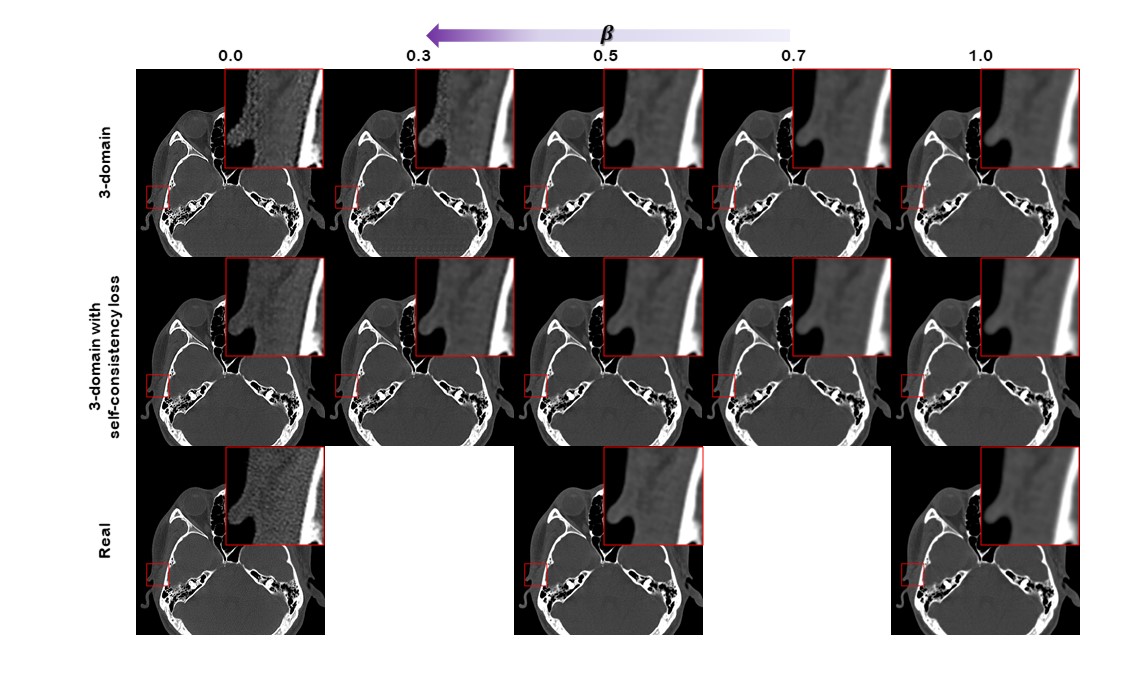}}
	\vspace*{-0.5cm}
	\caption{Ablation study results on self-consistency loss. The first row images are generated images without the self-consistency
	loss, and the second row images are generated with the self-consistency loss. The last row shows the real images. The right most real image is input, and the left and middle ones are target sharp and middle kernel images, respectively. }
	\label{fig_result_self_consistency}
\end{figure*}

The self-consistency loss was unique for three-domain learning of switchable cycleGAN (see \eqref{eq:loss_total}). As shown in Figure \ref{fig_result_self_consistency}, various artifacts appeared when training the model without this term.
Specifically, in the generation of sharp kernel images, artifacts occurred along the outlines of bones, 
 and irrelevant patterns appeared at the bottom of the generated images. However, these artifacts are no more present when using 
 the self-consistency loss.  This qualitative observation can be also supported by quantitative results in Table \ref{tab_result_3domain}. Compared to the model trained without self-consistency loss, all the PSNR and SSIM values were higher in the model trained with self-consistency loss.

\subsection{Autoencoder and Cycle-Consistency}

Note that the supervised learning should be able to preserve the fidelity of the structures since it has pixelwise constraint. 
By contrast, the unsupervised learning methods lacks this constraint. As a result, one may wonder whether our cycleGAN method may generate fake structures. However, our results show an opposite phenomenon. This interesting phenomenon is due to the use of identity-loss and cycle-consistency loss.
 As reported  in our prior work \cite{kang2019cycle},  when the cycle-consistency and identity losses were missing, spurious artifacts occured in the low-dose CT denoising, but  proper use of both cycle-consistency and identity loss prevented this from happening.

Similar to \cite{kang2019cycle}, in this paper we used both identity and cycle-consistency loss.
The identity loss, which is equal to the auto-encoder loss for the case of two-domain learning,
 plays a role to preserve the structure by providing pixel-wise constraint. 
 Additionally, the cycle-consistency loss in cycleGAN also poses strong pixel-wise constraint in that it ensures self-consistency when reverting to the original domain, which prevent from  fake structures being created. In contrast to the CycleGAN, the standard GAN lacks these two constraint, which leads to the creation of a falsified structure as reported in \cite{kang2019cycle}.

\subsection{Limitation}

Our method is not free of limitations. Though the proposed method could generate kernel images better than other methods and even comparable with the target images, fine details were not perfectly preserved in some cases such as trabecular bones. 
Also, as shown in Figure \ref{fig_result_alpha_fakeS}, the pixel value has increased around the brain-bone interface. For clinical applications such as cancer diagnosis, it is not a serious problem, but it could be misunderstood as bleeding in some applications. Although  one could confirm whether it is a true bleeding or an artifacts in this by adjusting $\beta$ values, this needs to be verified in real clinical scenario.
In addition, overshoot and undershoot around the bones prevented an accurate assessment of the thickness of the bones or nasal cavities (the first row of Figure \ref{fig_result_alpha_cartilage} and the third row of Figure \ref{fig_result_Polyphase}). This edge enhancement usually occurred when generating sharp kernel images using $\beta=0$. Though the edge enhancement problem can be relieved by using $\beta$ values greater than zero, more systematic study may be necessary.

\section{Conclusion}
\label{sec:conclusion}

Different properties of two kernels, sharp and soft kernels, produce two types of reconstructed images. Here, we have proposed a post-hoc image domain translation between these two kernels  to generate one kernel image from the other. Our proposed method was based on a switchable cycleGAN in combination with the adaptive instance normalization. Thanks to AdaIN, the conversion of kernels could be carried out with a single generator, and different images were generated along the optimal transport path by the combination of the two given kernels. 
Furthermore, intermediate domain kernel images can be effectively utilized  during the training by introducing split AdaIN code generators, which
significantly improved the feature domain interpolation performance.
The improved performance of our proposed model has been proven with extensive experimental results.


\end{document}